\documentclass[letterpaper, 10 pt, conference]{iros03}

\usepackage{times}

\providecommand{\tabularnewline}{\\}

\usepackage{epsfig}
\usepackage{amsfonts}
\usepackage{amsmath}
\usepackage{amssymb}
\usepackage{mathptmx}

\begin{document}

\title{\LARGE Robust Sound Source Localization Using a Microphone Array on
a Mobile Robot}

\author{\parbox{6 in}{\centering Jean-Marc Valin, Fran\c{c}ois Michaud, Jean Rouat, Dominic L\'{e}tourneau\\
LABORIUS - Research Laboratory on Mobile Robotics and Intelligent Systems\\
Department of Electrical Engineering and Computer Engineering\\
Universit\'{e} de Sherbrooke\\
Qu\'{e}bec, Canada\\
\texttt{jmvalin@jmvalin.ca}}
}

\maketitle\thispagestyle{empty}
\begin{abstract}
\footnotetext{\copyright 2003 IEEE.  Personal use of this material is permitted. Permission from IEEE must be obtained for all other uses, in any current or future media, including reprinting/republishing this material for advertising or promotional purposes, creating new collective works, for resale or redistribution to servers or lists, or reuse of any copyrighted component of this work in other works.}The hearing sense on a mobile robot is important because it is omnidirectional
and it does not require direct line-of-sight with the sound source.
Such capabilities can nicely complement vision to help localize a
person or an interesting event in the environment. To do so the robot
auditory system must be able to work in noisy, unknown and diverse
environmental conditions. In this paper we present a robust sound
source localization method in three-dimensional space using an array
of 8 microphones. The method is based on time delay of arrival estimation.
Results show that a mobile robot can localize in real time different
types of sound sources over a range of 3 meters and with a precision
of $3^{\circ}$.
\end{abstract}

\section{Introduction}

Compared to vision, robot audition is in its infancy: while research
activities on automatic speech recognition are very active, the use
and the adaptation of these techniques to the context of mobile robotics
has only been addressed by a few. There is the SAIL robot that uses
one microphone to develop online audio-driven behaviors \cite{zhang-weng2001}.
The robot ROBITA \cite{matsusaka-tojo-kubota-furukawa-tamiya-hayata-nakano-kobayashi99}
uses 2 microphones to follow a conversation between two persons. The
humanoid robot SIG \cite{nakadai-okuno-kitano2002,okuno-nakadai-kitano2002}
uses two pairs of microphones, one pair installed at the ear position
of the head to collect sound from the external world, and the other
placed inside the head to collect internal sounds (caused by motors)
for noise cancellation. Like humans, these last two robots use binaural
localization, i.e., the ability to locate the source of sound in three
dimensional space.

However, it is a difficult challenge to only use a pair of microphones
on a robot to match the hearing capabilities of humans. The human
hearing sense takes into account the acoustic shadow created by the
head and the reflections of the sound by the two ridges running along
the edges of the outer ears. With a pair of microphones, only localization
in two dimensions is possible, without being able to distinguish if
the sounds come from the front or the back of the robot. Also, it
may be difficult to get high precision readings when the sound source
is in the same axis of the pair of microphones.

It is not necessary to limit robots to a human-like auditory system
using only two microphones. Our strategy is to use more microphones
to compensate for the high level of complexity in the human auditory
system. This way, increased resolution can be obtained in three-dimensional
space. This also means increased robustness, since multiple signals
allow to filter out noise (instead of trying to isolate the noise
source by putting sensors inside the robot's head, as with SIG) and
discriminate multiple sound sources. It is with these potential benefits
in mind that we developed a sound source localization method based
on time delay of arrival (TDOA) estimation using an array of 8 microphones.
The method works for far-field and near-field sound sources and is
validated using a Pioneer 2 mobile robotic platform.

The paper is organized as follows. Section \ref{sec:TDOA-Estimation}
presents the principles behing TDOA estimation. Section \ref{sec:Position-estimation}
explains how the position of the sound source is derived from the
TDOA, followed by experimental results in Section \ref{sec:Experimental-setup}.

\section{TDOA Estimation\label{sec:TDOA-Estimation}}

We consider windowed frames of $N$ samples with 50\% overlap. For
the sake of simplicity, the index corresponding to the frame is ommitted
from the equations. In order to determine the delay in the signal
captured by two different microphones, it is necessary to define a
coherence measure. The most common coherence measure is a simple cross-correlation
between the signals perceived by two microphones, as expressed by:\begin{equation}
R_{ij}(\tau)=\sum_{n=0}^{N-1}x_{i}[n]x_{j}[n-\tau]\label{eq:TDOA_correlation_def}\end{equation}
where $x_{i}[n]$ is the signal received by microphone $i$ and $\tau$
is the correlation lag in samples. The cross-correlation $R_{ij}(\tau)$
is maximal when $\tau$ is equal to the offset between the two received
signals. The problem with computing the cross-correlation using Equation
\ref{eq:TDOA_correlation_def} is that the complexity is $\mathcal{O}\left(N^{2}\right)$.
However, it is possible to compute an approximation in the frequency
domain by computing the inverse Fourier transform of the cross-spectrum,
reducing the complexity to $\mathcal{O}\left(N\log_{2}N\right)$.
The correlation approximation is given by:\begin{equation}
R_{ij}(\tau)\approx\sum_{k=0}^{N-1}X_{i}(k)X_{j}(k)^{*}e^{\imath2\pi k\tau/N}\label{eq:TDOA_correlation_freq}\end{equation}
where $X_{i}(k)$ is the discrete Fourier transform of $x_{i}[n]$
and $X_{i}(k)X_{j}(k)^{*}$ is the cross-spectrum of $x_{i}[n]$ and
$x_{j}[n]$.

A major limitation of that approach is that the correlation is strongly
dependent on the statistical properties of the source signal. Since
most signals, including voice, are generally low-pass, the correlation
between adjacent samples is high and generates cross-correlation peaks
that can be very wide. 

The problem of wide cross-correlation peaks can be solved by whitening
the spectrum of the signals prior to computing the cross-correlation
\cite{Omologo}. The resulting {}``whitened cross-correlation''
is defined as:\begin{equation}
R_{ij}^{(w)}(\tau)=\sum_{k=0}^{N-1}\frac{X_{i}(k)X_{j}(k)^{*}}{\left|X_{i}(k)\right|\left|X_{j}(k)\right|}e^{\imath2\pi k\tau/N}\label{eq:TDOA_correlation_whiten}\end{equation}
and corresponds to the inverse Fourier transform of the normalized
(whitened) cross spectrum.

\begin{figure}[ht]
\begin{center}\includegraphics[%
  width=0.80\columnwidth,
  keepaspectratio]{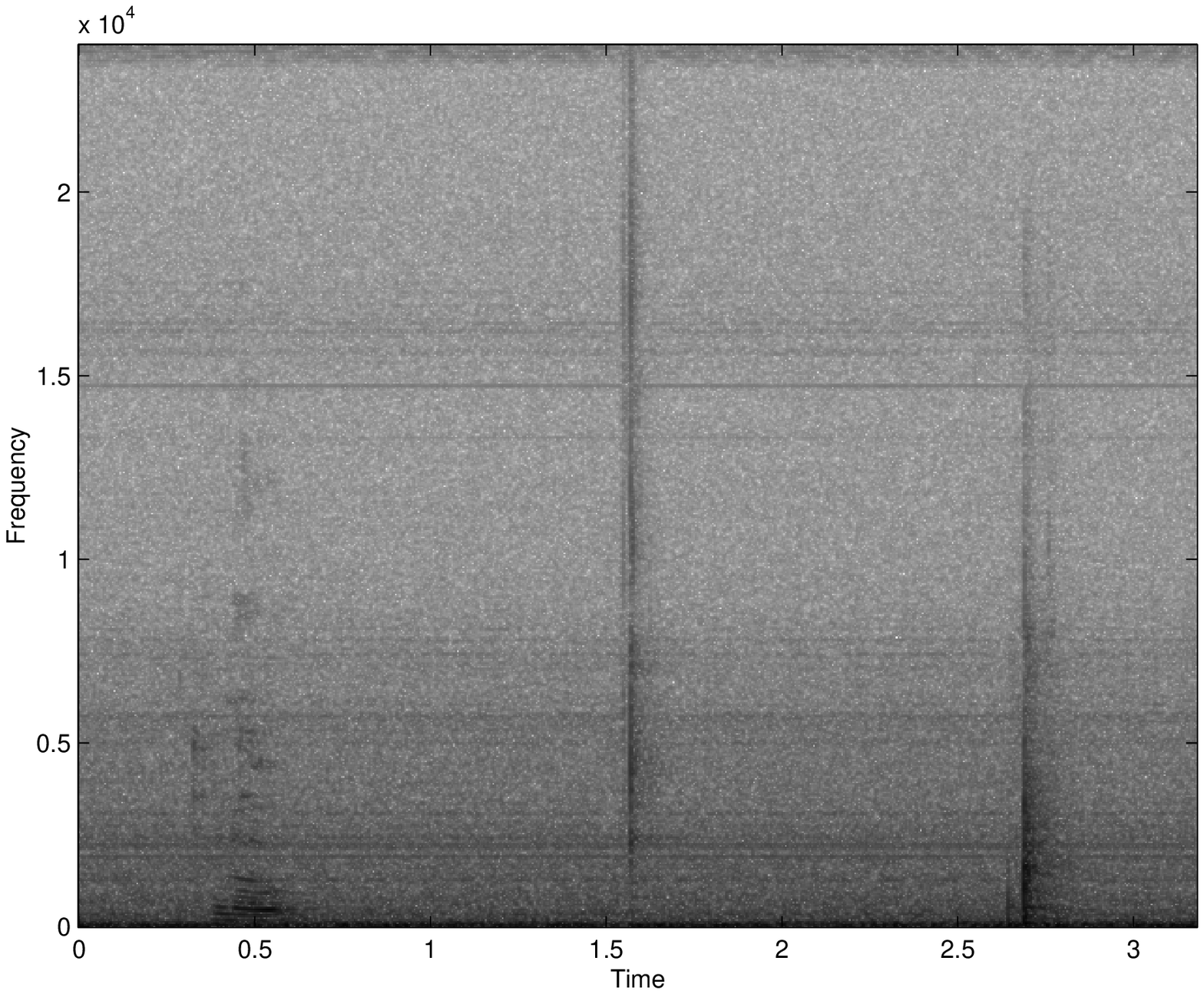}\end{center}

\caption{Spectrogram of the signal received at microphone 1 ($X_{1}(k)$)
for the following sounds: speech at 0.5 sec, finger snap at 1.5 sec
and boot noise on the floor at 2.7 sec.\label{cap:Spectrogram-mic1}}
\end{figure}

\begin{figure}[ht]
\begin{center}\includegraphics[%
  width=0.80\columnwidth,
  keepaspectratio]{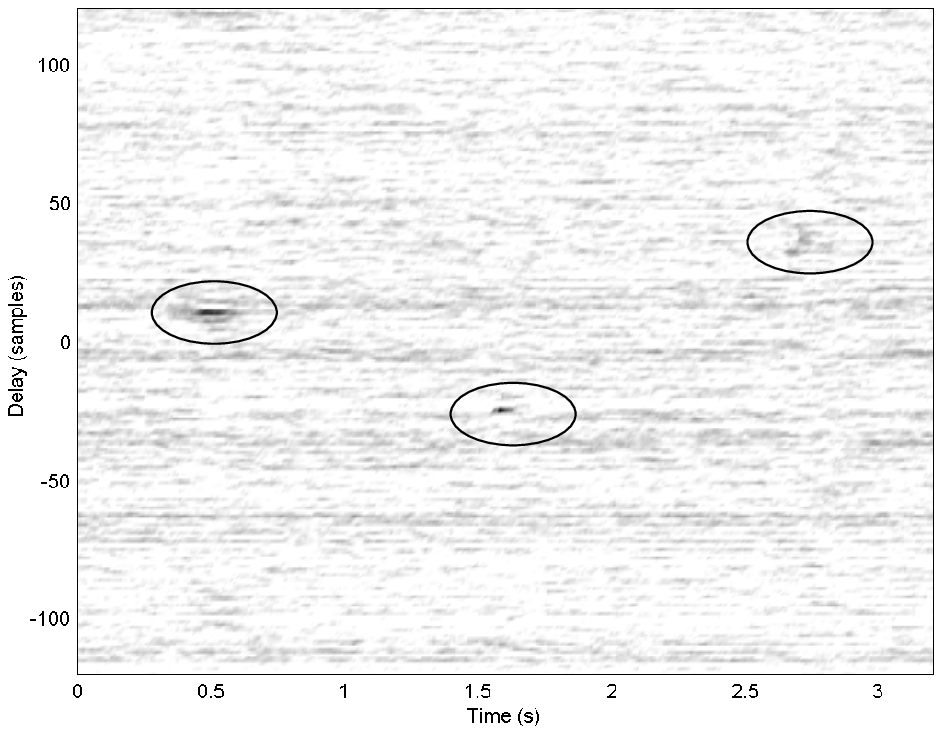}\end{center}

\caption{Whitened cross-correlation $R_{ij}^{(w)}(\tau)$ with peaks (circled)
corresponding to the sound sources.\label{cap:Whitened-cross-correlation}}
\end{figure}

Whitening allows to only take the phase of $X_{i}(k)$ into account,
giving each frequency component the same weight and narrowing the
wide maxima caused by correlations within the received signal. Figure
\ref{cap:Spectrogram-mic1} shows the spectrogram of the noisy signal
as received by one of the microphones in the array. The corresponding
whitened cross-correlation in Figure \ref{cap:Whitened-cross-correlation}
shows peaks at the same time as the sources found in Figure \ref{cap:Spectrogram-mic1}.

\subsection{Spectral Weighting}

The whitened cross-correlation method explained in the previous subsection
has several drawbacks. Each frequency bin of the spectrum contributes
the same amount to the final correlation, even if the signal at that
frequency is dominated by noise. This makes the system less robust
to noise, while making detection of voice (which has a narrow bandwidth)
more difficult.

In order to counter the problem, we developed a new weighting function
of the spectrum. This gives more weight to regions in the spectrum
where the local signal-to-noise ratio (SNR) is the highest. Let $X(k)$
be the mean power spectral density for all the microphones at a given
time and $X_{n}(k)$ be a noise estimate based on the time average
of previous $X(k)$. We define a noise masking weight by:\begin{equation}
w(k)=\max\left(0.1,\frac{X(k)-\alpha X_{n}(k)}{X(k)}\right)\label{eq:noise_weighting1}\end{equation}
where $\alpha<1$ is a coefficient that makes the noise estimate more
conservative. $w(k)$ becomes close to 0 in regions that are dominated
by noise, while $w(k)$ is close to 1 in regions where the signal
is much stronger than the noise. The second part of the weighting
function is designed to increase the contribution of tonal regions
of the spectrum (where the local SNR is very high). Starting from
Equation \ref{eq:noise_weighting1}, we define the enhanced weighting
function $w_{e}(k)$ as:\begin{equation}
w_{e}(k)=\left\{ \begin{array}{ll}
w(k) & ,\: X(k)\leq X_{n}(k)\\
w(k)\left(\frac{X(k)}{X_{n}(k)}\right)^{\gamma} & ,\: X(k)>X_{n}(k)\end{array}\right.\label{eq:noise_weighting2}\end{equation}
where the exponent $0<\gamma<1$ gives more weight to regions where
the signal is much higher than the noise. For our system, we empirically
set $\alpha$ to 0.4 and $\gamma$ to 0.3. The resulting weighted
cross-correlation is defined as:\begin{equation}
R_{ij}^{(e)}(\tau)=\sum_{k=0}^{N-1}\frac{w_{e}^{2}(k)X_{i}(k)X_{j}(k)^{*}}{\left|X_{i}(k)\right|\left|X_{j}(k)\right|}e^{\imath2\pi k\tau/N}\label{eq:TDOA_correlation_weighted}\end{equation}

\begin{figure}[ht]
\begin{center}\includegraphics[%
  width=0.80\columnwidth,
  keepaspectratio]{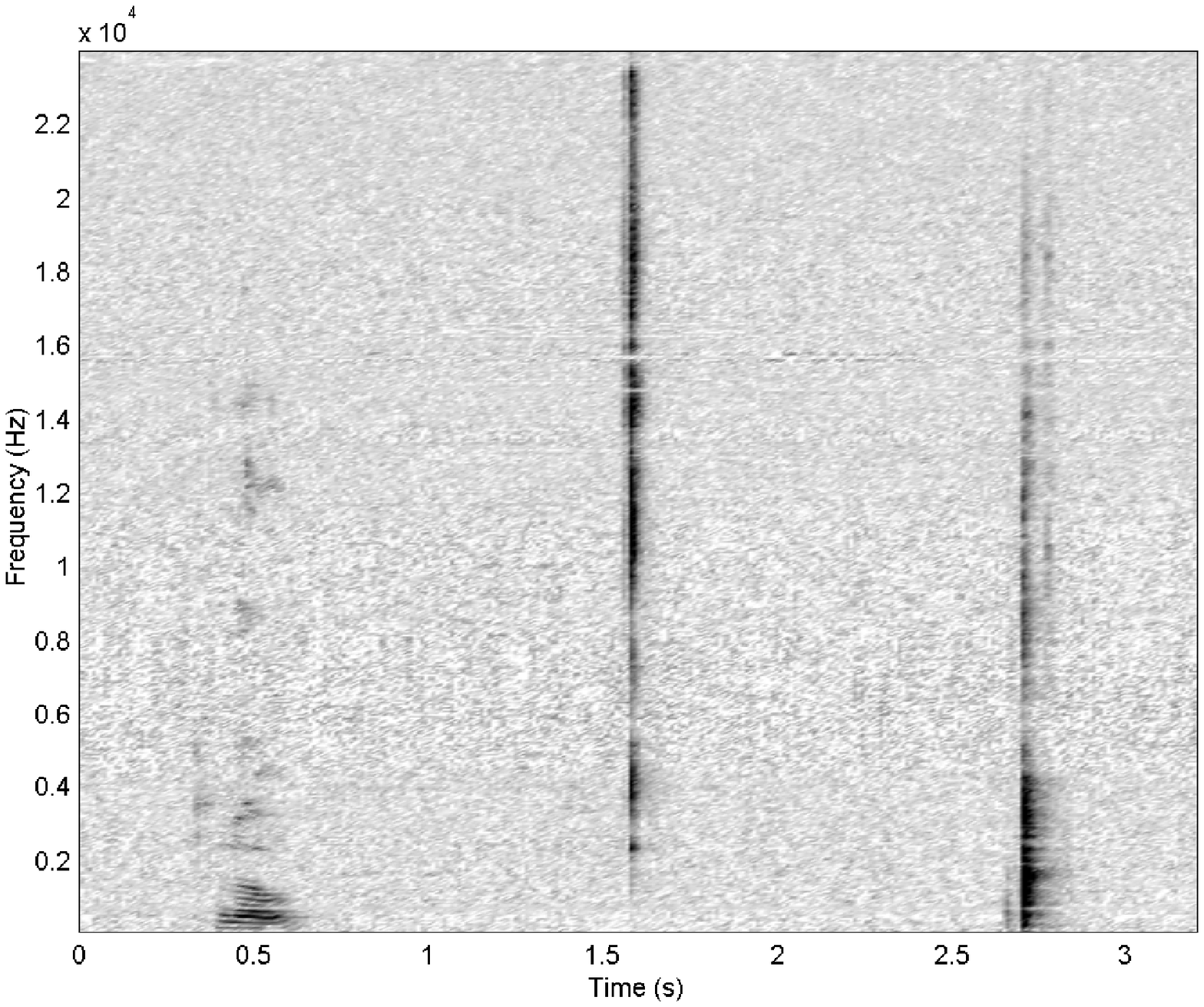}\end{center}

\caption{Noise weighting $w_{e}(k)$ for the sound sources\label{cap:Noise-weighting}}
\end{figure}

\begin{figure}[ht]
\begin{center}\includegraphics[%
  width=0.80\columnwidth,
  keepaspectratio]{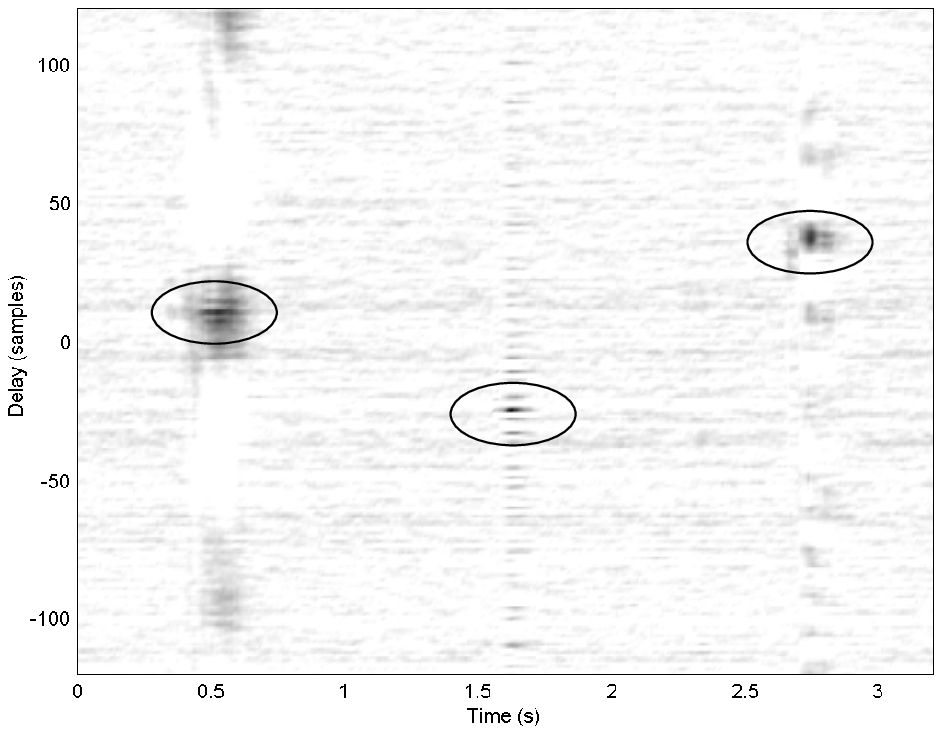}\end{center}

\caption{Cross-correlation with noise-weighting $R_{ij}^{(e)}(\tau)$ with
peaks (circled) corresponding to the sound sources.\label{cap:Cross-correlation-with-noise-weighting}}
\end{figure}

The value of $w_{e}(k)$ as a function of time is shown in Figure
\ref{cap:Noise-weighting} and the resulting cross-correlation is
shown in Figure \ref{cap:Cross-correlation-with-noise-weighting}.
Compared to Figure \ref{cap:Whitened-cross-correlation} it is possible
to see that the cross-correlation has less noise, although the peaks
are slightly wider. Nonetheless, the weighting method increases the
robustness of TDOA estimation.

\subsection{TDOA Estimation Using $N$ Microphones}

The time delay of arrival (TDOA) between microphones $i$ and $j$;
$\Delta T_{ij}$ can be found by locating the peak in the cross-correlation
as:\begin{equation}
\begin{array}{ccc}
\Delta T_{ij}= & \mathrm{argmax} & R_{ij}^{(e)}(\tau)\\
 & \tau\end{array}\label{eq:TDOA_argmax}\end{equation}

Using an array of $N$ microphones, it is possible to compute $N(N-1)/2$
different cross-correlations of which only $N-1$ are independent.
We chose to work only with the $\Delta T_{1i}$ values ($\Delta T_{12}$
to $\Delta T_{18}$), the remaining ones being derived by:\begin{equation}
\Delta T_{ij}=\Delta T_{1j}-\Delta T_{1i}\label{eq:delta_T_dependent}\end{equation}
The number of false detections can be reduced by considering sources
to be valid only when Equation \ref{eq:delta_T_dependent} is satisfied
for all $i\neq j$. 

Since in practice the highest peak may be caused by noise, we extract
the $M$ highest peaks in each cross-correlation (where $M$ is set
empirically to 8) and assume that one of them represents the real
value of $\Delta T_{ij}$. This leads to a search through all possible
combinations of $\Delta T_{1i}$ values (there are a total of $M^{N-1}$
combinations) that satisfy Equation \ref{eq:delta_T_dependent} for
all dependent $\Delta T_{ij}$. For example, in the case of an array
of 8 microphones, there are 7 independent delays ($\Delta T_{12}$
to $\Delta T_{18}$), but a total of 21 constraints (e.g. $\Delta T_{23}=\Delta T_{13}-\Delta T_{12}$),
which makes it very unlikely to falsely detect a source. When more
than one set of $\Delta T_{1i}$ values respect all the constraints,
only the one with the greatest correlation values is retained and
used to find the direction of the source using the method presented
in Section \ref{sec:Position-estimation}.

\section{Position Estimation\label{sec:Position-estimation}}

Once TDOA estimation is performed, it is possible to compute the position
of the source through geometrical calculations. One technique based
on a linear equation system \cite{Mahajan} but sometimes, depending
on the signals, the system is ill-conditioned and unstable. For that
reason, a simpler model based on far field assumption%
\footnote{It is assumed that the distance to the source is much larger than
the array aperture.%
} is used. 

\begin{figure}[ht]
\begin{center}\includegraphics[%
  scale=0.8]{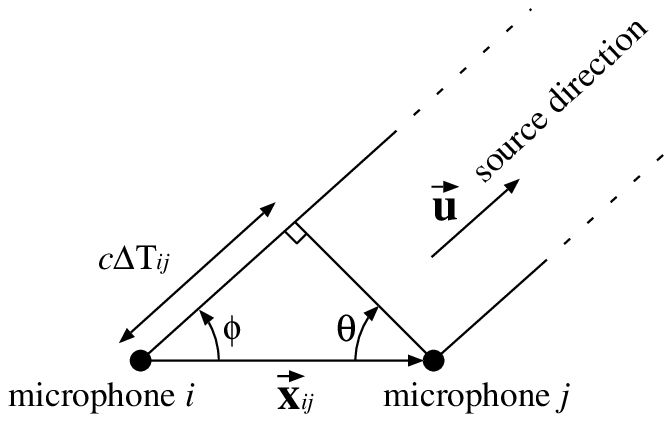}\end{center}

\caption{Computing source direction from TDOA.\label{cap:pos_geom}}
\end{figure}

Figure \ref{cap:pos_geom} illustrates the case of a 2 microphone
array with a source in the far-field. Using the cosine law, we can
state that:\begin{equation}
\cos\phi=\frac{\vec{\mathbf{u}}\cdot\vec{\mathbf{x}}_{ij}}{\left\Vert \vec{\mathbf{u}}\right\Vert \left\Vert \vec{\mathbf{x}}_{ij}\right\Vert }=\frac{\vec{\mathbf{u}}\cdot\vec{\mathbf{x}}_{ij}}{\left\Vert \vec{\mathbf{x}}_{ij}\right\Vert }\label{eq:cosine_law}\end{equation}
where $\vec{\mathbf{x}}_{ij}$ is the vector that goes from microphone
$i$ to microphone $j$ and $\vec{\mathbf{u}}$ is a unit vector pointing
in the direction of the source. From the same figure, it can be stated
that:\begin{equation}
\cos\phi=\sin\theta=\frac{c\Delta T_{ij}}{\left\Vert \vec{\mathbf{x}}_{ij}\right\Vert }\label{eq:deltaT_sin_theta}\end{equation}
where $c$ is the speed of sound. When combining equations \ref{eq:cosine_law}
and \ref{eq:deltaT_sin_theta}, we obtain:\begin{equation}
\vec{\mathbf{u}}\cdot\vec{\mathbf{x}}_{ij}=c\Delta T_{ij}\label{eq:dot-prod_equals_TDOA}\end{equation}
which can be re-written as:\begin{equation}
u\left(x_{j}-x_{i}\right)+v\left(y_{j}-y_{i}\right)+w\left(z_{j}-z_{i}\right)=c\Delta T_{ij}\label{eq:dot-prod_equals_TDOA2}\end{equation}
where $\vec{\mathbf{u}}=(u,\: v,\: w)$ and $\vec{\mathbf{x}}_{ij}=\left(x_{j}-x_{i},\: y_{j}-y_{i},\: z_{j}-z_{i}\right)$,
the position of microphone $i$ being $\left(x_{i},\: y_{i},\: z_{i}\right)$.
Considering $N$ microphones, we obtain a system of $N-1$ equations: 

\begin{equation}
\begin{array}{c}
\left[\begin{array}{ccc}
\left(x_{2}-x_{1}\right) & \left(y_{2}-y_{1}\right) & \left(z_{2}-z_{1}\right)\\
\left(x_{3}-x_{1}\right) & \left(y_{2}-y_{1}\right) & \left(z_{3}-z_{1}\right)\\
\vdots & \vdots & \vdots\\
\left(x_{N}-x_{1}\right) & \left(y_{N}-y_{1}\right) & \left(z_{N}-z_{1}\right)\end{array}\right]\left[\begin{array}{c}
u\\
v\\
w\end{array}\right]\\
=\left[\begin{array}{c}
c\Delta T_{12}\\
c\Delta T_{13}\\
\vdots\\
c\Delta T_{1N}\end{array}\right]\end{array}\label{eq:pos_linear_system}\end{equation}

In the case with more than 4 microphones, the system is over-constrained
and the solution can be found using the pseudo-inverse, which can
be computed only once since the matrix is constant. Also, the system
is guaranteed to be stable (i.e., the matrix is non-singular) as long
as the microphones are not all in the same plane.

\begin{figure}[ht]
\begin{center}\includegraphics[%
  width=0.90\columnwidth,
  keepaspectratio]{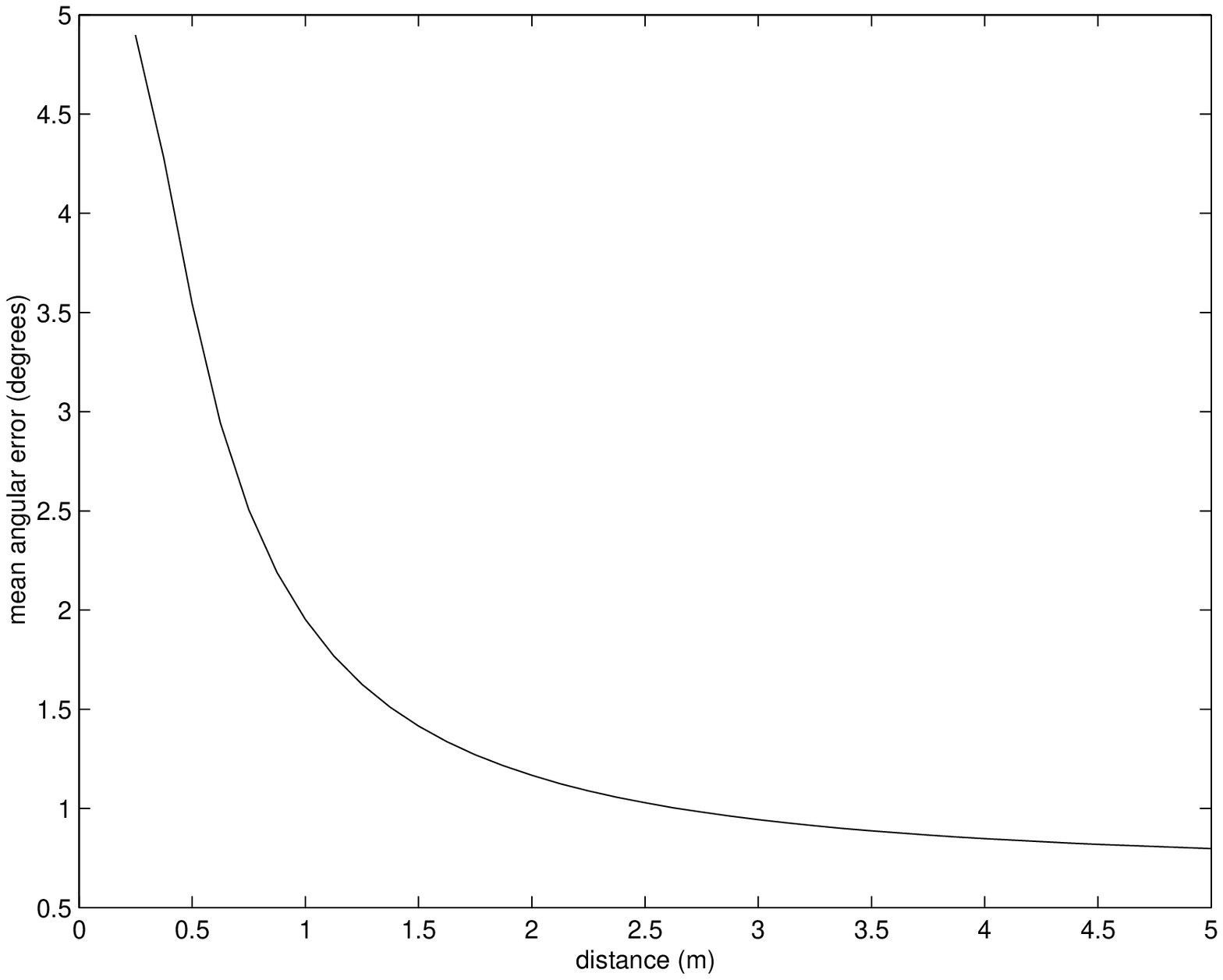}\end{center}

\caption{Mean angular error as a function of distance between the sound source
and the center of the array for near-field.\label{cap:nearfield}}
\end{figure}

The linear system expressed by Relation \ref{eq:pos_linear_system}
is theoretically valid only for the far-field case. In the near field
case, the main effect on the result is that the direction vector $\vec{\mathbf{u}}$
found has a norm smaller than unity. By normalizing $\vec{\mathbf{u}}$,
it is possible to obtain results for the near field that are almost
as good as for the far field. Simulating an array of 50 cm $\times$
40 cm $\times$ 36 cm shows that the mean angular error is reasonable
even when the source is very close to the array, as shown by Figure
\ref{cap:nearfield}. Even at 25 cm from the center of the array,
the mean angular error is only 5 degrees. At such distance, the error
corresponds to about 2-3 cm, which is often larger than the source
itself. For those reasons, we consider that the method is valid for
both near-field and far-field. Normalizing $\vec{\mathbf{u}}$ also
makes the system insensitive to the speed of sound because Equation
\ref{eq:pos_linear_system} shows that $c$ only has an effect on
the magnitude of $\vec{\mathbf{u}}$. That way, it is not necessary
to take into account the variations in the speed of sound.

\section{Results\label{sec:Experimental-setup}}

\begin{figure}[ht]
\begin{center}\includegraphics[%
  width=0.80\columnwidth,
  keepaspectratio]{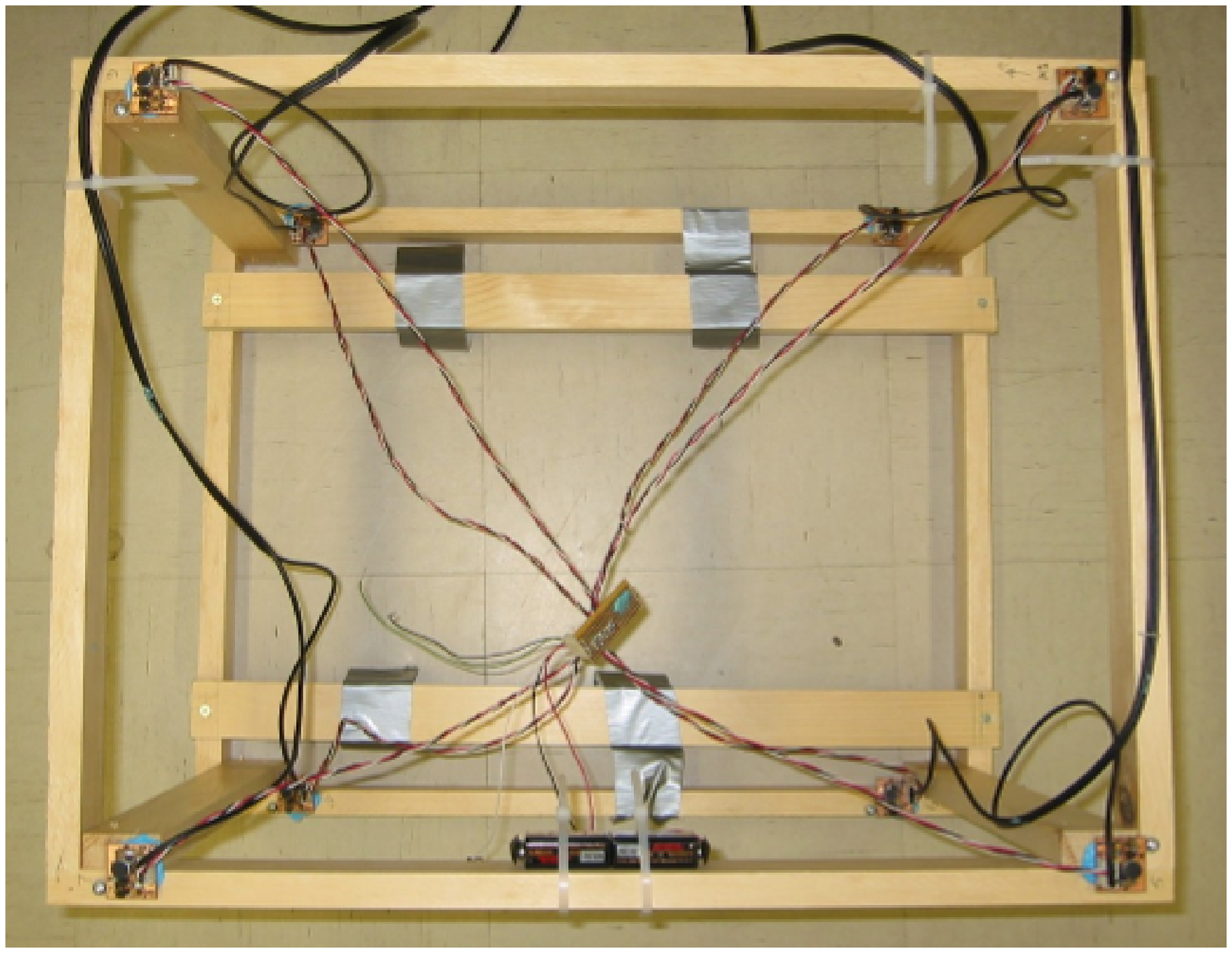}\end{center}

\caption{Top view of an array of 8 microphones mounted on a rectangular prism
of dimensions 50 cm $\times$ 40 cm $\times$ 36 cm.\label{cap:pic_array}}
\end{figure}

\begin{figure}[ht]
\begin{center}\includegraphics[%
  width=0.80\columnwidth,
  keepaspectratio]{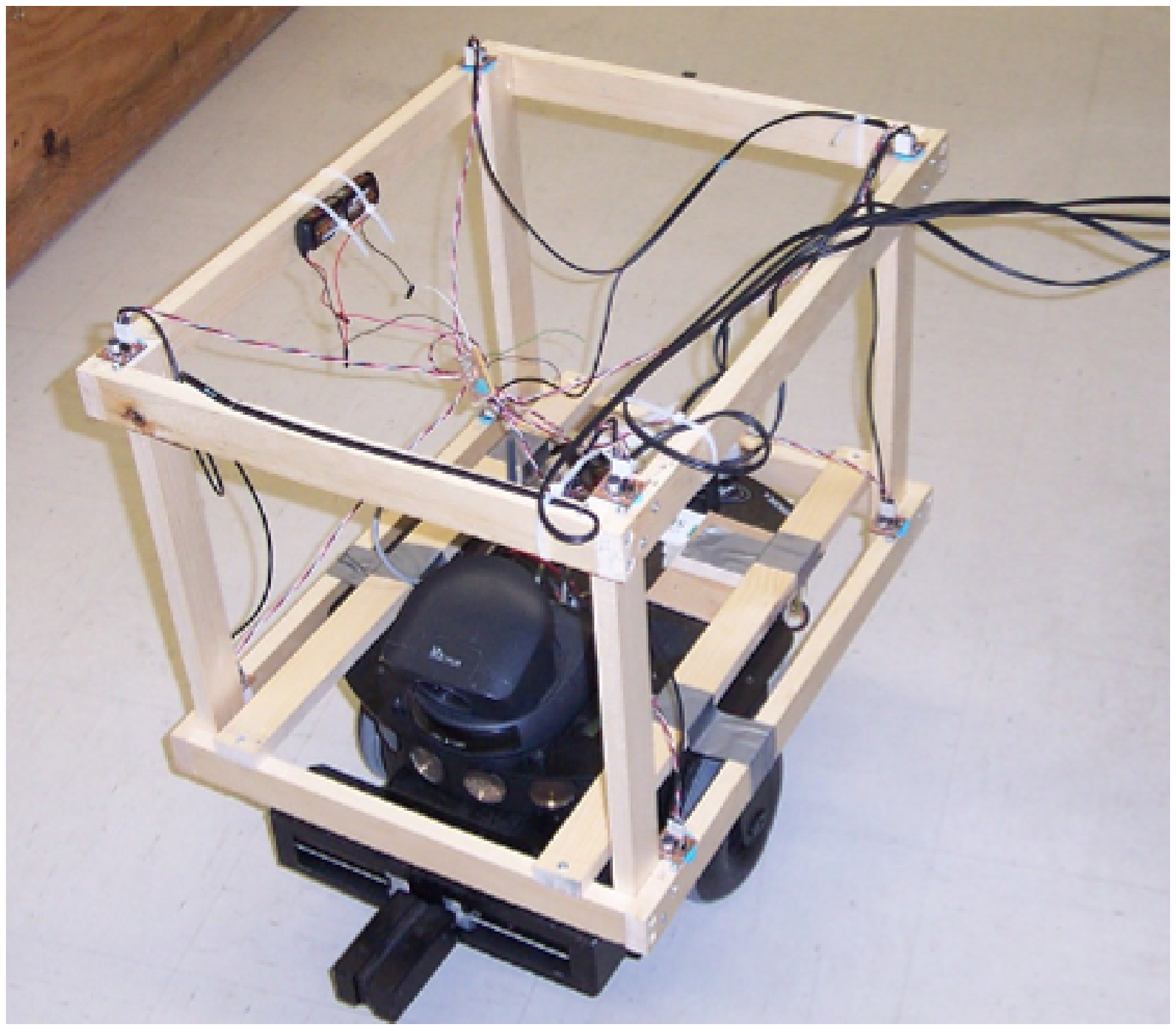}\end{center}

\caption{Microphone array installed on a Pioneer 2 robot.\label{cap:array-pioneer2}}
\end{figure}

The array used for experimentation is composed of 8 microphones arranged
on the summits of a rectangular prism, as shown in Figure \ref{cap:pic_array}.
The array is mounted on an ActivMedia Pioneer 2 robot, as shown in
Figure \ref{cap:array-pioneer2}. However, due to processor and space
limitations (the acquisition is performed using an 8-channel PCI soundcard
that cannot be installed on the robot), the signal acquisition and
processing is performed on a desktop computer (Athlon XP 2000+). The
algorithm described requires about 15\% CPU to work in real-time. 

\begin{figure}[ht]
\begin{center}\includegraphics[%
  width=0.80\columnwidth,
  keepaspectratio]{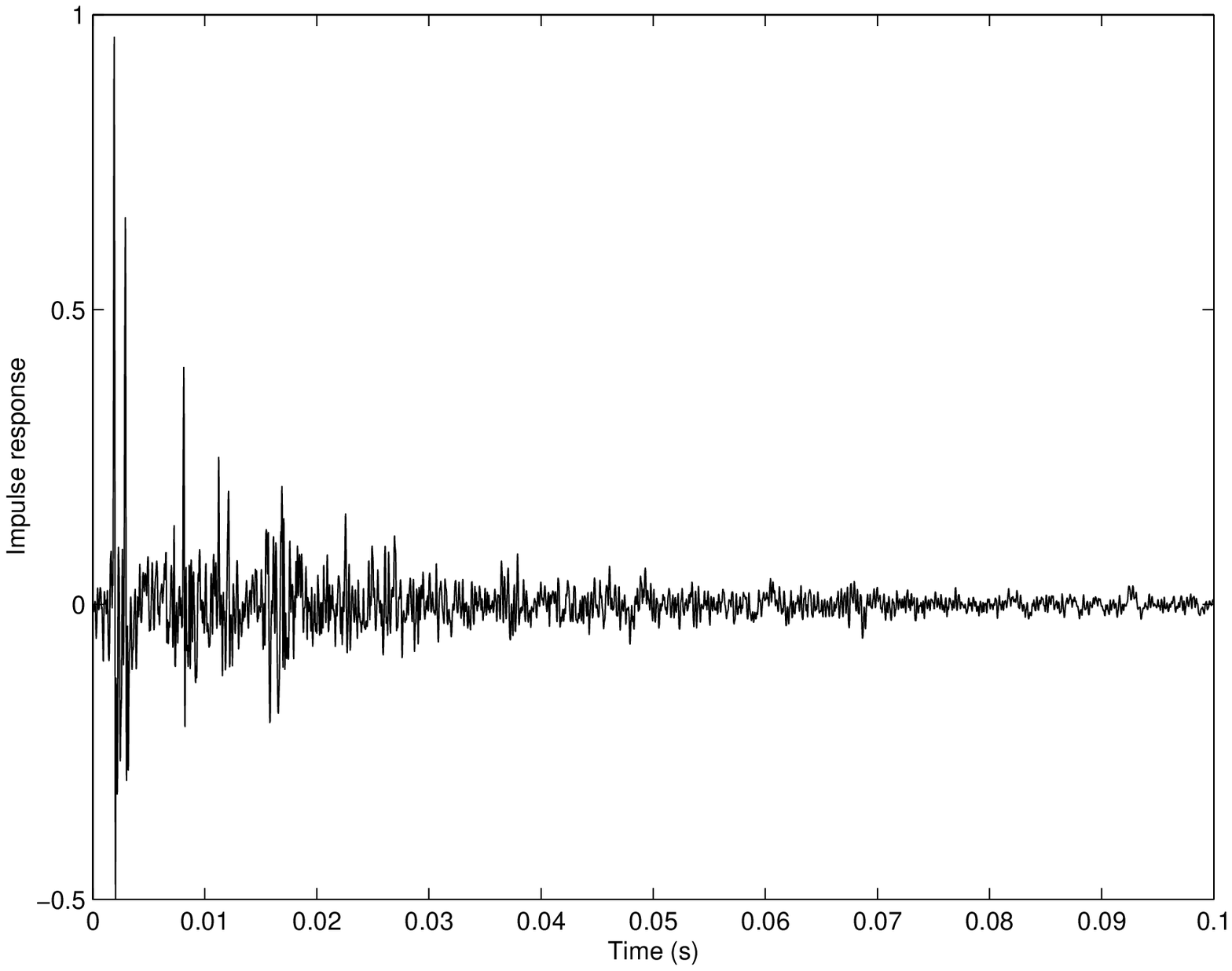}\end{center}

\caption{Impulse response of room reverberation. Secondary peaks represent
reflections on the floor and on the walls.\label{cap:reverb_impulse_response}}
\end{figure}

The localization system mounted on a Pioneer 2 is used to direct the
robot's camera toward sound sources. The horizontal angle is used
to rotate the robot in the source direction, while the vertical angle
is used to control the tilt of the camera. The system is evaluated
in a room with a relatively high noise level (as shown from the spectrogram
in Figure \ref{cap:Spectrogram-mic1}), mostly due to several fans
in proximity. The reverberation is moderate and its corresponding
transfer function is shown in Figure \ref{cap:reverb_impulse_response}.

The system was tested with sources placed in different locations in
the environment. In each case, the distance and elevation are fixed
and measures are taken for different horizontal angles. The mean angular
error for each configuration is shown in Table \ref{cap:Measured-localization-error}.
It is worth mentioning that part of this error, mostly at short distance,
is due to the difficulty of accurately positionning the source and
to the fact that the speaker used is not a point source. Other sources
of error come from reverberation on the floor (more important when
the source is high) and from the near-field approximation as shown
in Figure \ref{cap:nearfield}. Overall, the angular error is the
same regardless of the direction in the horizontal plane and varies
only slightly with the elevation, due to the interference from floor
reflections. This is an advantage over systems based on two microphones
where the error is high when the source is located on the sides \cite{nakadai-okuno-kitano2002}.

\begin{table}[ht]
\caption{Measured mean angular localization error\label{cap:Measured-localization-error}}
\begin{center}\begin{tabular}{|c|c|}
\hline 
Distance, Elevation&
Mean Angular Error\tabularnewline
\hline
\hline 
3 m, $-7^{\circ}$&
$1.7^{\circ}$\tabularnewline
\hline 
3 m, $8^{\circ}$&
$3.0^{\circ}$\tabularnewline
\hline 
1,5 m, $-13^{\circ}$&
$3.1^{\circ}$\tabularnewline
\hline 
0,9 m, $24^{\circ}$&
$3.3^{\circ}$\tabularnewline
\hline
\end{tabular}\end{center}

\end{table}

\begin{figure}[ht]
\begin{center}\includegraphics[%
  width=0.50\columnwidth,
  keepaspectratio]{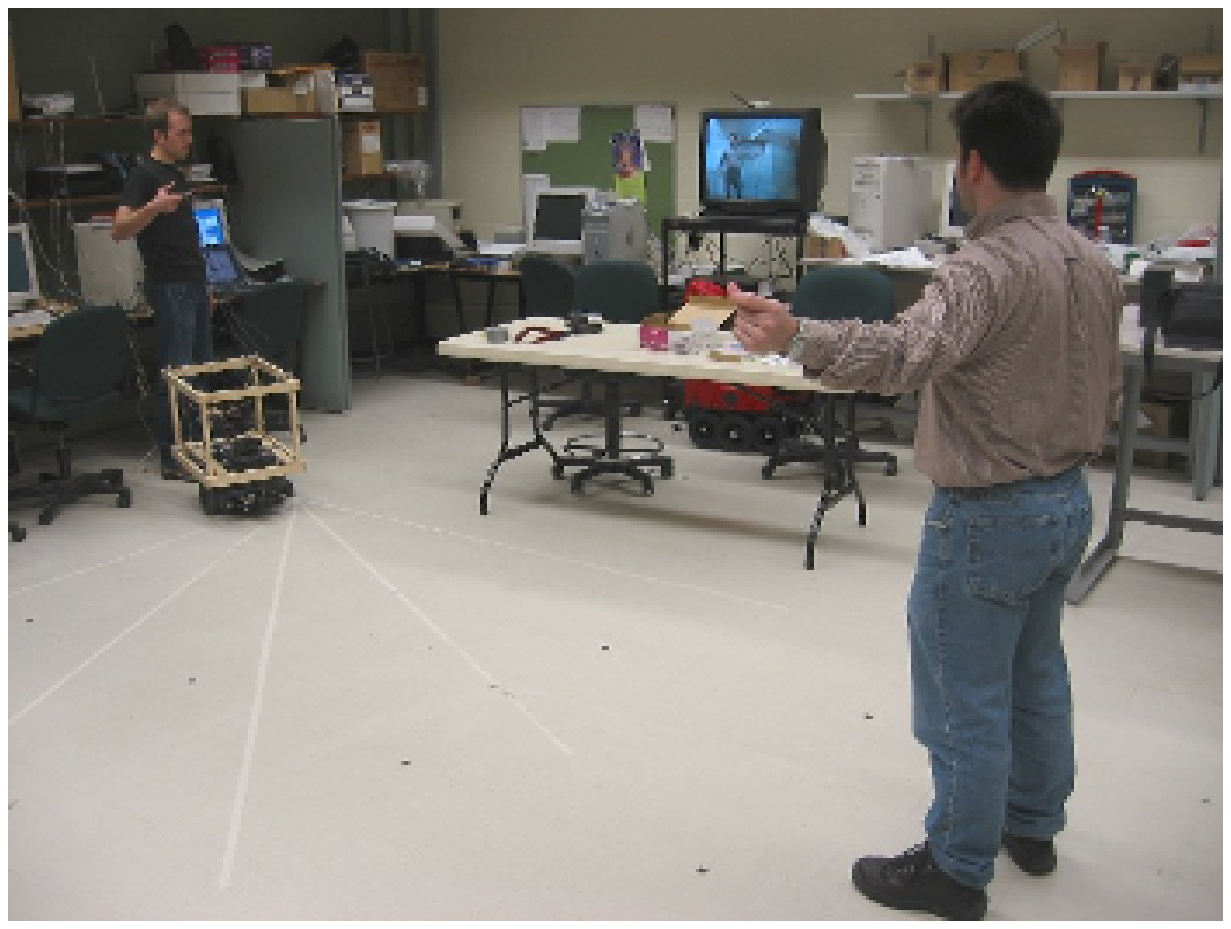}\includegraphics[%
  width=0.50\columnwidth,
  keepaspectratio]{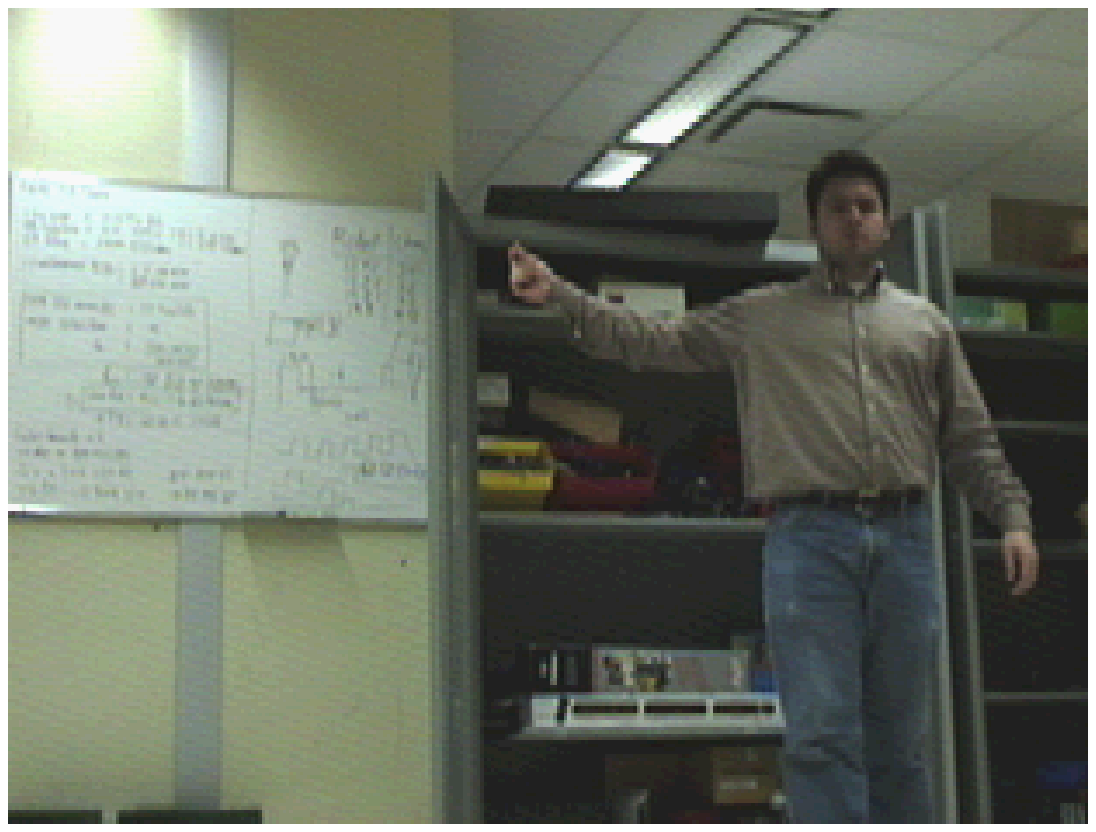}\end{center}\begin{center}(a) \hspace{4cm}(b)\end{center}

\begin{center}\includegraphics[%
  width=0.50\columnwidth,
  keepaspectratio]{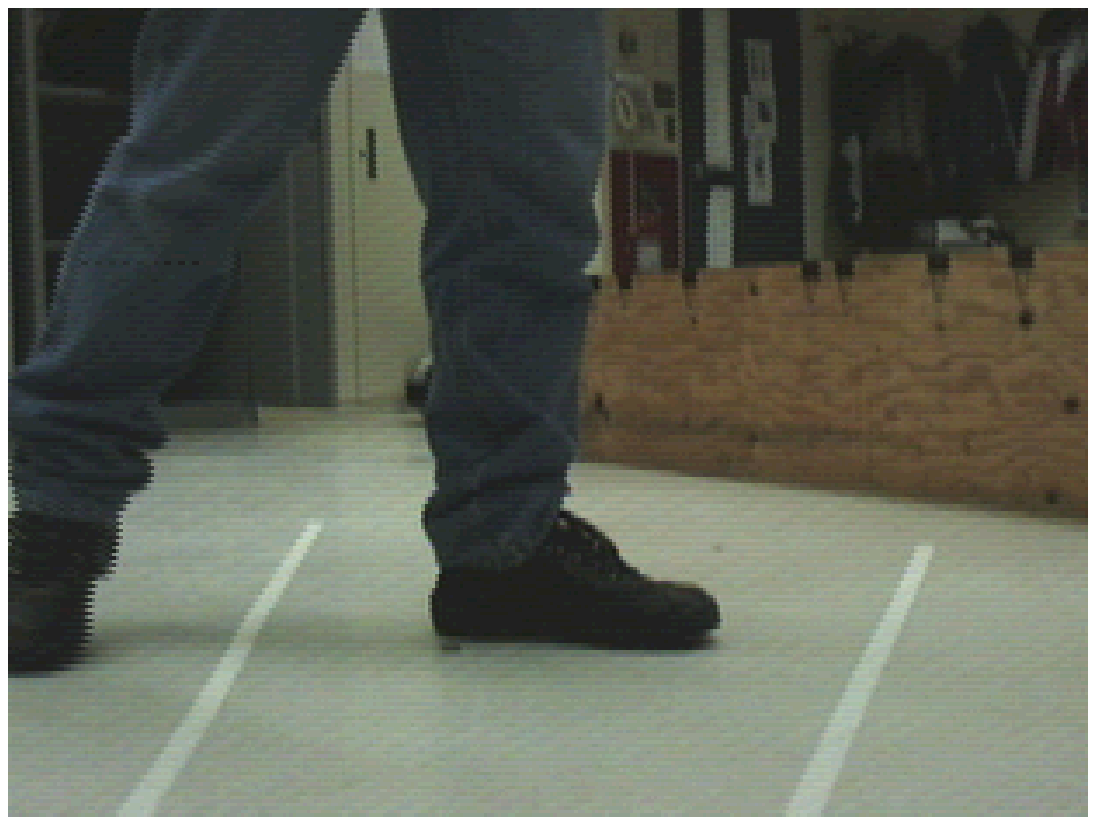}\includegraphics[%
  width=0.50\columnwidth,
  keepaspectratio]{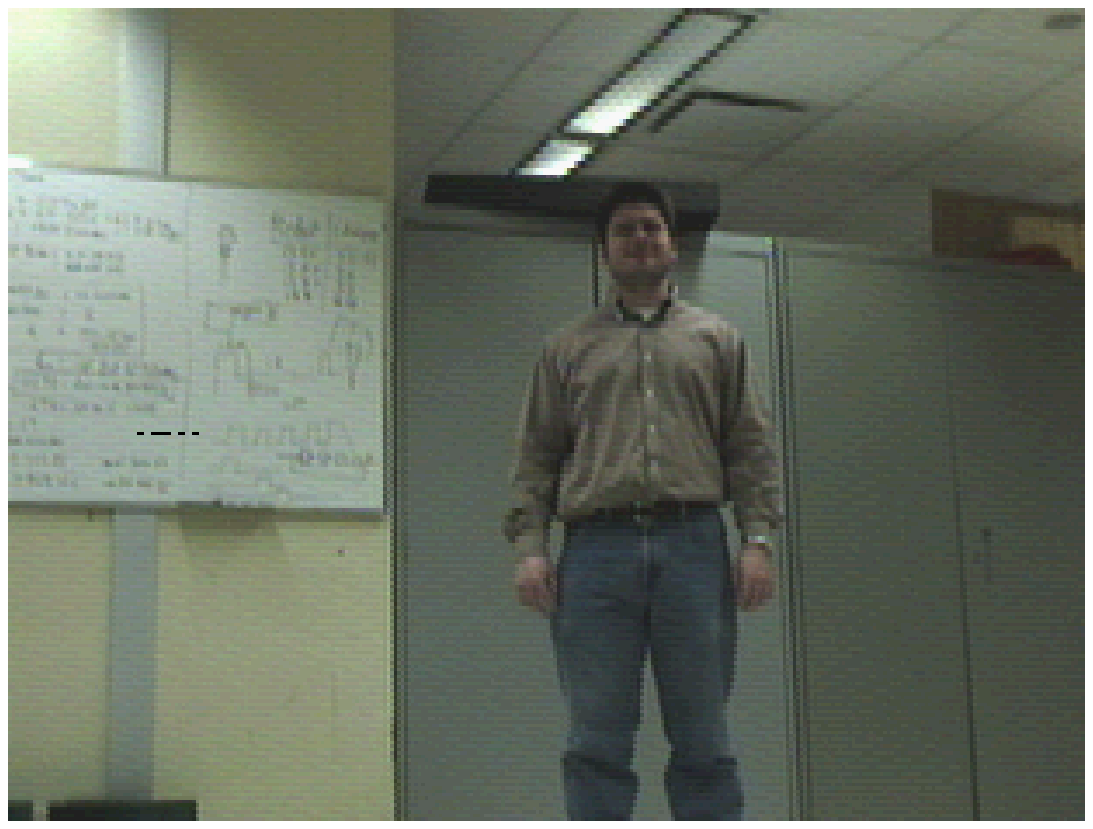}\end{center}\begin{center}(c) \hspace{4cm}(d)\end{center}

\caption{Photographs taken during experimentation. a) Experimental setup.
b) Snapping fingers at a distance of $\sim5\:\mathrm{m}$. c) Tapping
foot at $\sim2\:\mathrm{m}$. d) Speaking at a distance of $\sim5\:\mathrm{m}$.\label{cap:Photograph-taken-by}}
\end{figure}

Unlike other works where the localization is performed actively during
positioning \cite{Nakadai}, our approach is to localize the source
before even moving the robot, which means that the source does not
have to be continuous. In order to achieve that, the sound source
localization system is disabled while the robot is moving toward the
source. During a conversation between two or more persons, the robot
alternates between the talkers. In presence of two simultaneous sources,
the dominant one is naturally selected by the localization system.
Figure \ref{cap:Photograph-taken-by} shows the experimental setup
and images from the robot camera after localizing a source and moving
its camera toward it. Most of the positionning error in the image
is due to various actuator inaccuracies and the fact that the camera
is not located exactly at the center of the array.

Experiments show that the system functions properly up to a distance
between 3 and 5 meters, though this limitation is mainly a result
of the noise and reverberation conditions in the laboratory. Also,
not all sound types are equally well detected. Because of the whitening
process explained in Section \ref{sec:TDOA-Estimation}, each frequency
has roughly the same importance in the cross-correlation. This means
that when the sound to be localized is a tone, only a very small region
of the spectrum contains useful information for the localization.
The cross-correlation is then dominated by noise. This makes tones
very hard to localize using the current algorithm. We also observed
that this localization difficulty is present at a lesser degree for
the human auditory system, which cannot accurately localize sinusoids
in space.

On the other hand, some sounds are very easily detected by the system.
Most of these sounds have a large bandwidth, like fricatives, fingers
snapping, paper shuffling and percussive noises (object falling, hammer).
For voice, the detection usually happens within the first two syllables.

\section{Conclusion}

Using an array of 8 microphones, we have implemented a system that
accurately localizes sounds in three dimensions. Moreover, our system
is able to perform localization even on short-duration sounds and
does not require the use of any noise cancellation method. The precision
of the localization is $3^{\circ}$ over 3 meters. 

The TDOA estimation used in the system is shown to be relatively robust
to noise and reverberation. Also, the algorithm for transforming the
TDOA values to a direction is stable and independent of the speed
of sound.

In its current form, the presented system still lacks some functionality.
First, it cannot estimate the source distance. However, early simulations
indicate that it would be possible to estimate the distance up to
approximately 2 meters. Also, though possible in theory, the system
is not yet capable of localizing two or more simultaneous sources
and only the dominant one is perceived. In the case of more than one
speaker, the {}``dominant sound source'' alternates and it is possible
to estimate the direction of both speakers.

\newpage
\section*{Acknowledgment}

Fran\c{c}ois Michaud holds the Canada Research Chair (CRC) in Mobile
Robotics and Autonomous Intelligent Systems. This research is supported
financially by the CRC Program, the Natural Sciences and Engineering
Research Council of Canada (NSERC) and the Canadian Foundation for
Innovation (CFI). Special thanks to Serge Caron and Nicolas B\'{e}gin
for their help in this work. 

\bibliographystyle{plain}
\bibliography{iros,BiblioAudible}

\end{document}